# Rapid tracking through strongly scattering media with physics-informed neuromorphic speckle analysis

Yuqing Cao, Shuo Zhu, Rongzhou Chen, Jingyan Chen, Ni Chen and Edmund Y. Lam[*]

Department of Electrical and Computer Engineering, The University of Hong Kong, Pokfulam, Hong Kong, China.

*Corresponding author(s). E-mail(s): elam@eee.hku.hk;

**Abstract**

This work addresses the critical problem of tracking fast-moving objects through strongly scattering media in a low-light environment. Different from existing approaches that use frame-based cameras with fixed exposure times, which trade off signal-to-noise ratio for temporal resolution, we introduce computational neuromorphic tracking (CNT), a physics-informed framework that combines asynchronous event sensing with task-driven speckle analysis for robust motion estimation. We formulate the neuromorphic speckle aggregation as a spatiotemporal speckle representation, jointly optimizing the temporal and spatial parameters to maximize tracking stability under extreme conditions. Extensive experiments demonstrate that our method enables robust motion tracking of **10×** faster motion and under **10×** dimmer illumination compared to conventional systems. These improvements significantly broaden the operational regime for tracking through scattering media, providing an efficient and scalable solution for demanding scenarios involving rapid motion and low-light conditions.

# 1 Introduction

Motion tracking through strongly scattering media is foundational for many applications, including remote sensing and biomedical monitoring [1, 2]. Since ballistic photons are rapidly attenuated and multiple scattering severely distorts spatial information [3, 4], the detected signals consist predominantly of randomized speckle [2]. As a result, direct image formation becomes unreliable or even infeasible [6, 7]. This raises a fundamental question: whether and how motion can be reliably estimated from seemingly random speckle measurements when clear images are unavailable [8-10], particularly under extreme conditions such as rapid motions and low photon budgets [11, 12].

Existing speckle-based approaches, such as speckle correlography, attempt to answer this question by estimating motion from temporal correlations of speckle patterns [13, 14]. While computationally efficient, their performance is constrained by both measurement conditions and underlying physics [15, 16]. In particular, reliable correlation requires that speckle structure remains observable within the optical memory effect (OME), the small angular or lateral range over which speckle patterns maintain correlation [17]. Typically, these methods rely on conventional frame-based cameras, which suffer from an inherent trade-off: longer exposure improves signal-to-noise ratio (SNR) but introduces motion blur, whereas shorter exposure preserves temporal detail but reduces photon counts [18]. This trade-off degrades speckle contrast and temporal coherence under high-speed or low-light conditions [19, 20], precisely the regimes where robust tracking is most needed. As a result, the reliability of frame-based speckle correlography deteriorates significantly when objects are moving faster or when illumination levels drop.

Neuromorphic vision sensors offer an alternative acquisition paradigm by asynchronously reporting per-pixel brightness changes with microsecond latency and high dynamic range [21-24]. This decouples temporal resolution from exposure time and enables flexible temporal aggregation after acquisition [25, 26]. In principle, such sensors can alleviate the exposure-speed trade-off inherent to frame-based imaging. However, replacing the sensing modality alone does not resolve the tracking problem. Raw event streams are sparse and noise-sensitive, particularly under low-light conditions, and do not directly yield stable speckle representations suitable for correlation analysis [20]. The central challenge, therefore, lies not only in acquisition, but also in constructing a computational representation that preserves both temporal dynamics and speckle structure from asynchronous measurements.

Recent efforts explore data-driven reconstruction or learning-based inference from speckle or neuromorphic measurements [7, 10], while others investigate direct processing of event streams [20, 27]. These methods demonstrate the potential of asynchronous sensing, but often rely on implicit representations or heuristic parameter choices. They do not explicitly address the conditions required for stable speckle correlation under fast motion and photon-limited regimes. In particular, how to jointly preserve temporal coherence and speckle structure for reliable speckle-based tracking remains insufficiently resolved.

Consequently, we pioneer computational neuromorphic tracking (CNT), a joint hardware-algorithm framework for motion estimation through strongly scattering media. CNT encodes hidden motion clues into asynchronous event streams and decodes them via physics-informed speckle analysis. The advantage of CNT over conventional frame-based correlography is its ability to handle rapid motion without blur or loss of signal under low-light conditions. Additionally, CNT produces more stable, noise-resilient speckle maps than event-only methods that use raw streams directly. These advantages enable calibration-free tracking without prior knowledge of the medium or motion model. Experimentally, CNT increases the operational range for both object speed and illumination by $10\times$ compared to frame-based tracking. It achieves reliable continuous tracking where other modalities fail, particularly in high-speed and low-light scenarios.

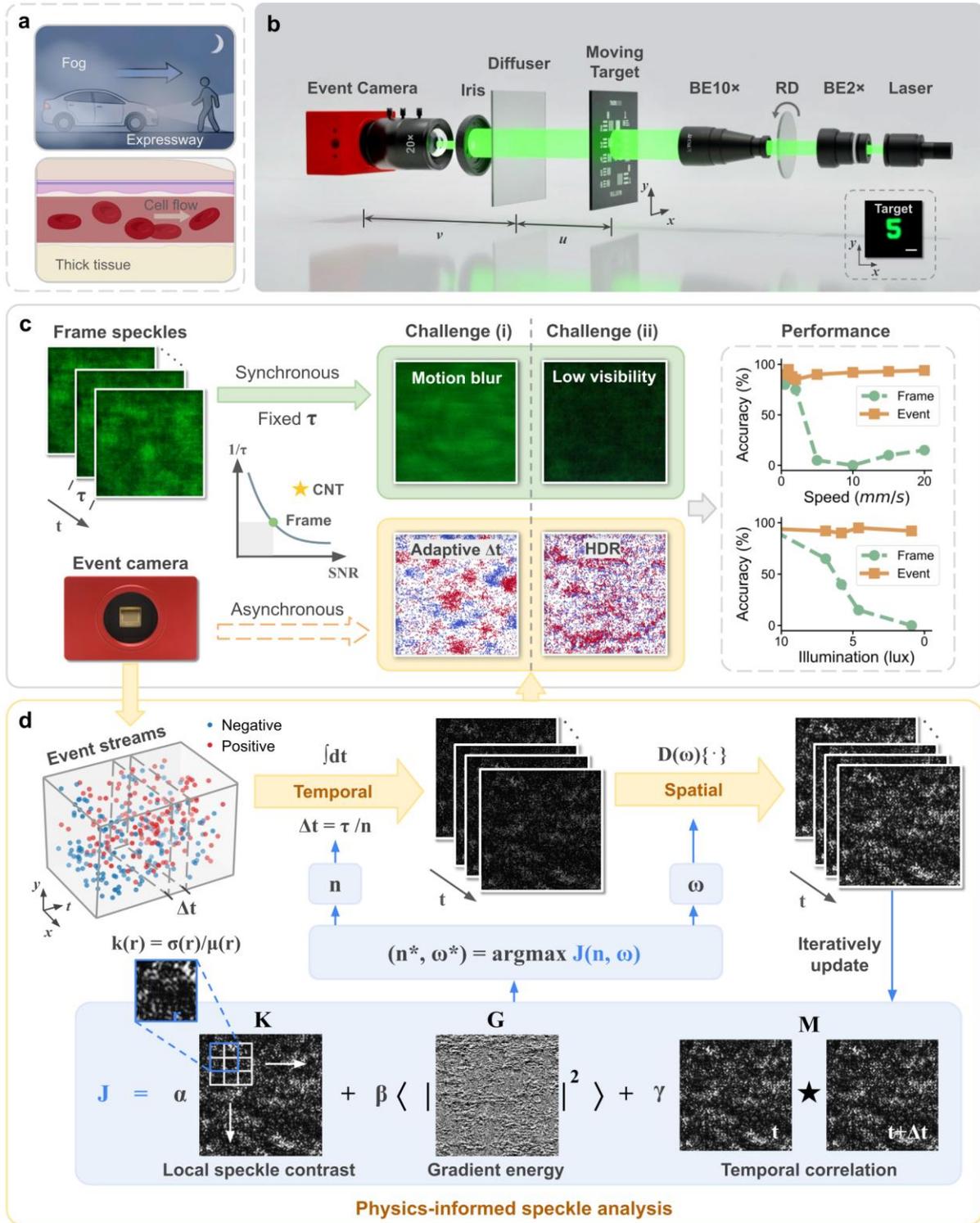

**Fig. 1 Overview of the CNT framework.** (a) Representative application scenarios: autonomous driving in fog at night (top) and noninvasive cell monitoring through biological tissue (bottom), both characterized by challenges (i) rapid motion and (ii) low photon flux. (b) Schematic of the CNT hardware configuration. (c) Comparison of imaging paradigms. Frame-based systems rely on synchronous exposure, coupling temporal resolution and SNR, whereas CNT uses asynchronous neuromorphic sensing to extend the accessible range of velocity and illumination. (d) Workflow of physics-informed speckle analysis. Event streams are aggregated into neuromorphic speckle maps governed by the spatiotemporal parameter pair $(n, \omega)$, which are jointly selected via a physics-informed objective function. Scale bar: 2 mm.

## 2 Results

### 2.1 CNT framework

Figure 1 presents an overview of the CNT framework. As illustrated in Figure 1a, scenarios such as autonomous navigation in fog and non-invasive monitoring through biological tissue involve fast motion and low photon flux. CNT is designed to address both challenges to enable future applications to such scenarios. To experimentally validate CNT, we built the optical setup shown in Figure 1b. Using this system, we compare CNT against conventional methods. As summarized in Figure 1c, CNT reliably tracks motion under high-speed and low-light conditions where frame-based methods fail due to motion blur or signal loss. This performance advantage stems from the physics-informed speckle analysis illustrated in Figure 1d.

CNT constructs a speckle representation parameterized by a temporal ratio $n$ and a spatial cutoff frequency $\omega$. Here $n$ balances temporal decorrelation against event statistics, while $\omega$ suppresses noise-dominated high-frequency components that destabilize correlation. Rather than relying on heuristic parameter selection, CNT optimizes a composite objective that directly measures motion observability:

$$\mathcal{J}(n, \omega) = \alpha K(n, \omega) + \beta G(n, \omega) + \gamma M(n, \omega), \tag{1}$$

where $\alpha$, $\beta$, and $\gamma$ are weights chosen empirically and fixed throughout all experiments. $K$ denotes local speckle contrast, $G$ spatial gradient energy, and $M$ temporal correlation [2]. These terms capture complementary requirements for stable correlation: contrast ensures signal visibility, spatial gradients preserve structural detail, and temporal correlation maintains consistency across measurements. The optimal parameters are obtained as

$$(n^*, \omega^*) = \underset{n, \omega}{\mathrm{argmax}}\, \mathcal{J}(n, \omega). \tag{2}$$

With these optimized parameters, CNT forms the final speckle maps and then recovers motion via speckle correlography. Under the optical memory effect (OME), the correlation peak position remains invariant to the medium's transmission properties [2], enabling calibration-free tracking without prior knowledge of the medium. CNT explicitly accounts for the OME constraint through adaptive temporal sampling. Let $d_{\mathrm{OME}}$ denote the maximum displacement that preserves speckle correlation [6]. For a reference exposure $\tau$, the displacement per aggregation window is $v\tau/n$. By selecting $n$ such that $v\tau/n < d_{\mathrm{OME}}$, CNT ensures that each measurement remains within the correlation-preserving regime. Recursive reference updates then enable tracking over larger cumulative displacements.

Together, CNT integrates asynchronous sensing with physics-informed representation design to maintain motion observability under fast motion and low-light conditions. The following subsections experimentally validate this capability.

## 2.2 CNT validation

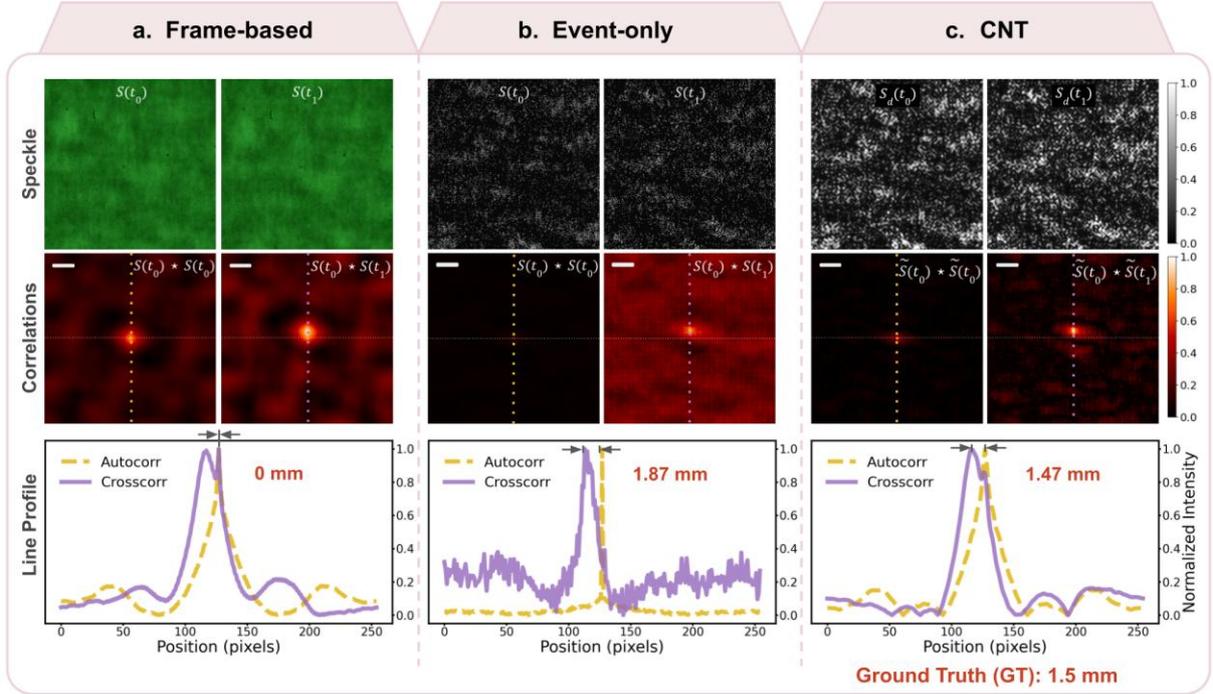

**Fig. 2  Experimental comparison of CNT with frame-based and event-only approaches.** An object translates at 1.5 mm/s. Correlations (middle row) are computed from successive speckle maps between one second (top row). Line profiles (bottom row) show autocorrelation (dashed) and cross-correlation (solid). (a) Frame-based imaging shows no resolvable peak shift, resulting in failed displacement estimation. (b) Event-only aggregation produces noisy correlations and a biased estimate of 1.87 mm. (c) CNT yields a clear correlation peak shift with an estimated displacement of 1.47 mm. Scale bars: 4 mm.

We evaluate how different sensing modalities and speckle analysis strategies affect displacement estimation. Data are collected using the setup in Figure 1b, with object speed fixed at $v = 1.5$ mm/s and ground truth displacement 1.5 mm. For CNT, the optimized parameters are $n = 1$ and $\omega = 50$. Figure 2 compares correlation behavior across methods. Frame-based speckle patterns (Figure 2a) are blurred and low contrast due to temporal integration, yielding a correlation peak at the origin and an estimated displacement of 0 mm. The event-only method (Figure 2b) produces grainy speckle maps with strong side-lobe noise, resulting in a shifted peak at 1.87 mm. This error arises from stochastic event clusters that generate spurious correlation maxima under low SNR conditions. In contrast, CNT (Figure 2c) produces spatiotemporally consistent speckle maps that preserve structural contrast while suppressing noise-dominated components. The resulting correlation peak is sharp and well-defined, yielding an estimated displacement of 1.47 mm. Quantitatively, CNT achieves a relative error of 2.22%, compared to 24.4% for the event-only method and 100% for the frame-based method. These results show that accurate tracking requires not only appropriate sensing but also a representation that jointly preserves temporal dynamics and speckle structure.

## 2.3 Effect of temporal parameter

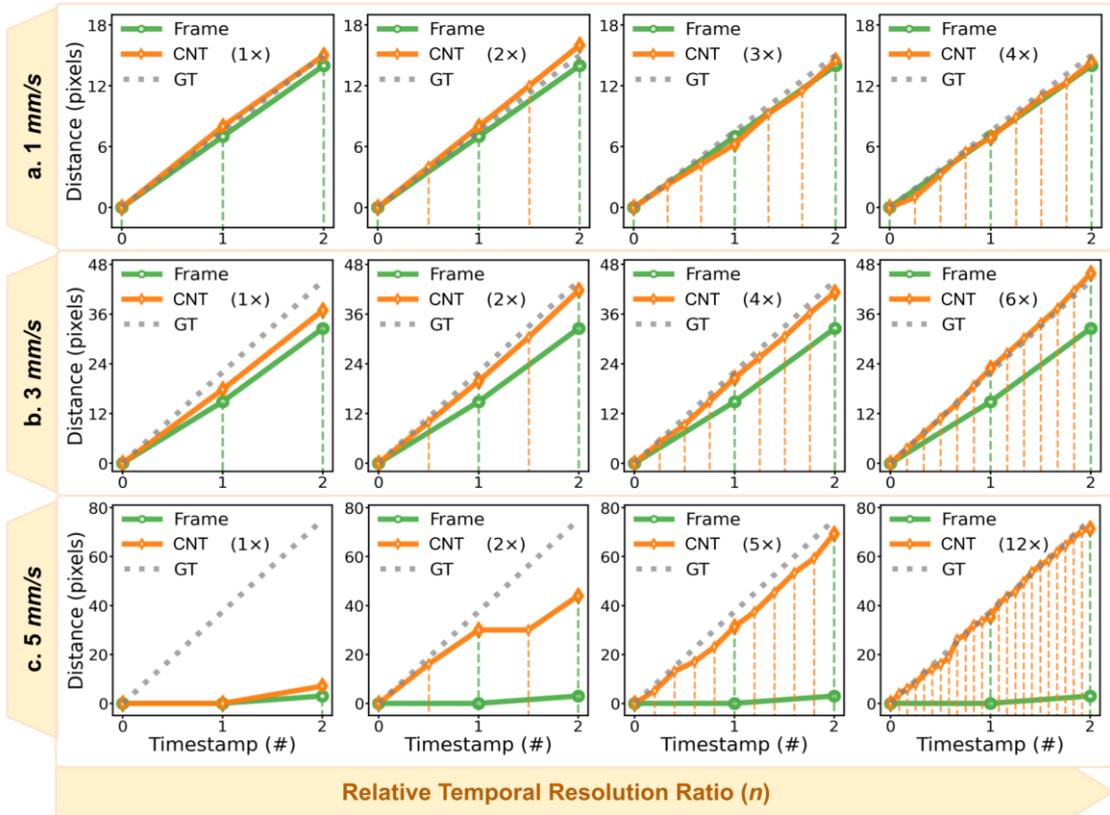

**Fig. 3  Effect of the temporal parameter $n$ on tracking performance.** Distance versus time for target speeds of (a) 1 mm/s, (b) 3 mm/s, and (c) 5 mm/s. Each panel compares ground truth (dotted), frame-based tracking (green), and CNT (orange) under different values of $n$. CNT performance depends strongly on $n$.

We investigate the dependence of tracking performance on the temporal ratio $n = \tau/\Delta t$, where $\Delta t$ is the event aggregation window. Figure 3 shows displacement error versus $n$ for different motion speeds. A clear trade-off is observed. Small $n$ (long aggregation windows) leads to temporal blurring and reduced sensitivity to fast dynamics, whereas large $n$ (short windows) produces sparse event statistics and unstable correlation peaks. The optimal $n$ shifts with motion speed: faster motion requires finer temporal sampling to remain within the memory effect, while slower motion benefits from longer integration to improve SNR.

Suboptimal choices of $n$ lead to either temporal decorrelation or insufficient event statistics, both of which degrade correlation stability. Another experiment conducted on a different object is included in Supplementary Note 1, leading to a similar observation.

## 2.4 Joint optimization of spatiotemporal parameters

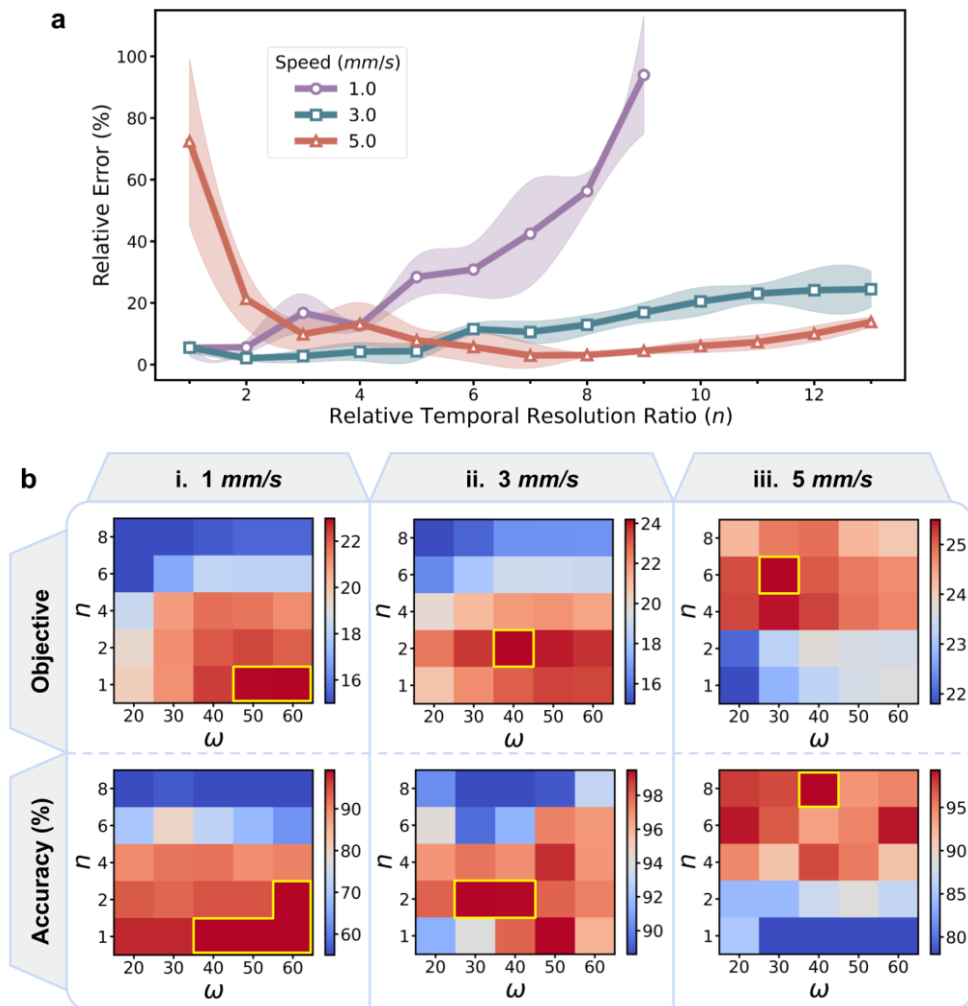

**Fig. 4 Joint optimization of spatiotemporal parameters.** (a) Relative tracking error as a function of temporal resolution ratio $n$ for different motion speeds. Each speed exhibits a distinct optimal $n$ that minimizes tracking error. (b) Joint two-dimensional optimization over spatiotemporal parameter pair $(n, \omega)$. Top row: value of the constructed objective function. Bottom row: measured tracking accuracy (%). Yellow-highlighted regions indicate optimal parameter pairs. The objective-function maxima closely align with the empirical accuracy maxima.

We evaluate the objective function $\mathcal{J}(n, \omega)$ as a joint function of temporal and spatial parameters. Tracking accuracy is measured using relative speed error. Figure 4a shows error versus $n$ at different speeds, each exhibiting a distinct minimum, indicating that each motion regime requires a specific temporal scale. Figure 4b presents a two-dimensional analysis over $(n, \omega)$, where the maxima of $\mathcal{J}$ closely coincide with the minima of tracking error. This agreement indicates that $\mathcal{J}$ captures key factors governing speckle correlation stability, namely contrast, structural preservation, and temporal consistency. This provides a criterion for selecting spatiotemporal parameters without heuristic tuning.

## 2.5 Speed boundary

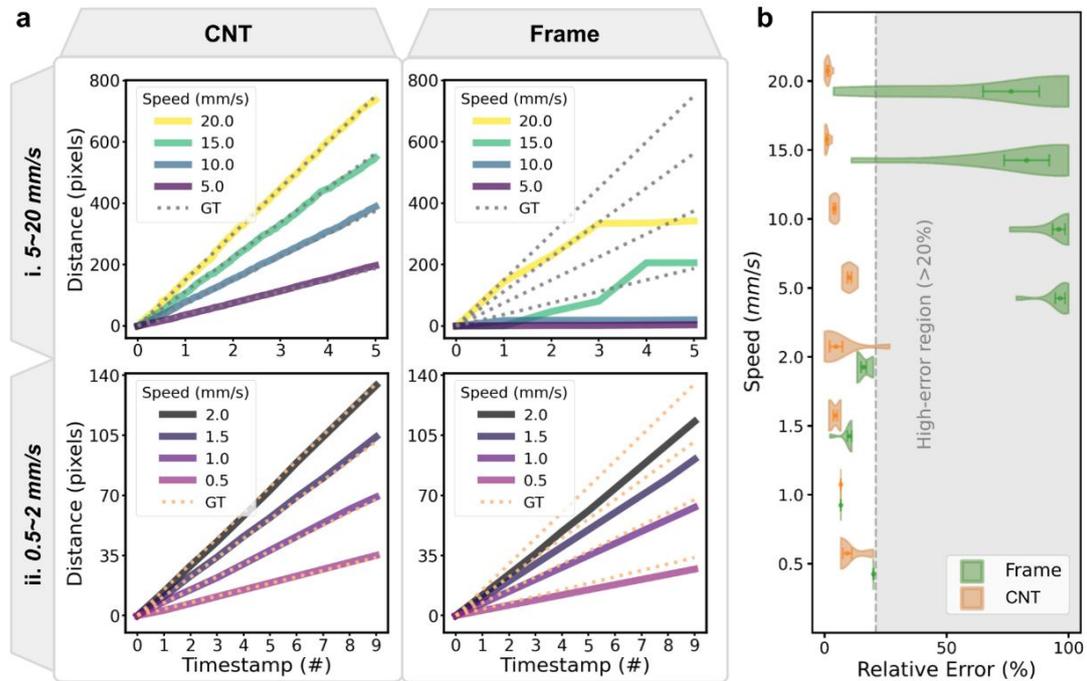

**Fig. 5  Cross-speed generalizability assessment.** (a) Displacement over time for target speeds from 0.5 to 20.0 mm/s, separated into (i) super-boundary and (ii) sub-boundary regimes relative to the single-pixel blur limit under 25 fps. CNT (left column) follows the ground truth (dotted) across all speeds, while frame-based tracking (right column) fails at super-boundary regime. (b) Violin plots of relative tracking error versus speed for frame-based (green) and CNT (orange) methods. The shaded region indicates errors exceeding 20%.

We evaluate tracking performance relative to the single-pixel motion blur limit of a frame-based system at 25 fps, corresponding to approximately 3.3 mm/s (see Supplementary Note 2). Figure 5a shows displacement trajectory. Below this limit, both methods follow the ground truth. Above it, frame-based tracking saturates and underestimates displacement due to motion-induced decorrelation, while CNT maintains linear trajectories across all tested speeds. Figure 5b shows the corresponding error distributions. Frame-based errors increase sharply beyond the blur limit, frequently exceeding 50% with large variance. In contrast, CNT maintains low error and narrow distributions up to 20 mm/s. Supplementary Note 3 compares the performance of the CNT and frame-based method on different directions with the same velocity, where CNT is more robust across various directions. These results indicate that fixed-exposure imaging imposes a practical speed limit, while CNT extends the accessible regime through asynchronous sensing and adaptive temporal representation.

## 2.6 Illumination boundary

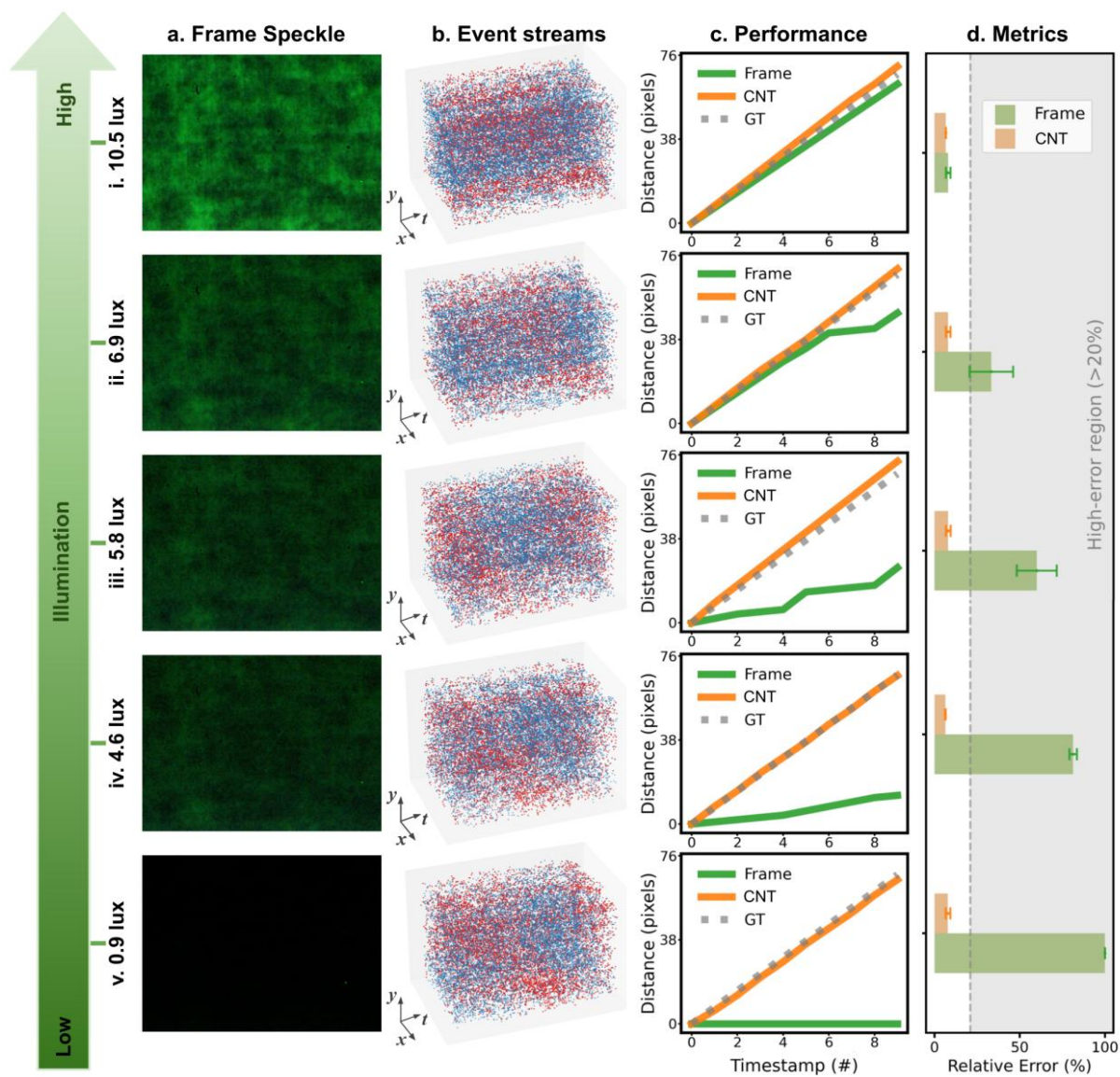

**Fig. 6 Cross-illumination generalizability assessment.** (a) Speckle frames acquired at 25 fps under decreasing illumination. (b) Corresponding event streams over 40 ms. (c) Cumulative displacement estimated by CNT (orange) and frame-based (green) methods compared with ground truth (dotted). (d) Relative tracking error at the final time point. The shaded region indicates a high error region where errors exceed 20%.

We evaluate performance under photon-limited conditions from 10.5 lux to 0.9 lux. As illumination decreases, frame-based speckle patterns lose contrast and spatial detail (Figure 6a), while event streams retain activity (Figure 6b). Quantitative results (Figure 6c, d) show that frame-based error increases rapidly, reaching 100% at 0.9 lux. CNT maintains errors below 10% across the full range. This degradation reflects the exposure-limited trade-off between SNR and motion blur in frame-based systems. CNT avoids this constraint by preserving temporal information through asynchronous sampling and optimizing the spatiotemporal representation, enabling stable tracking under low-light conditions.

## 2.7 Trajectory reconstruction

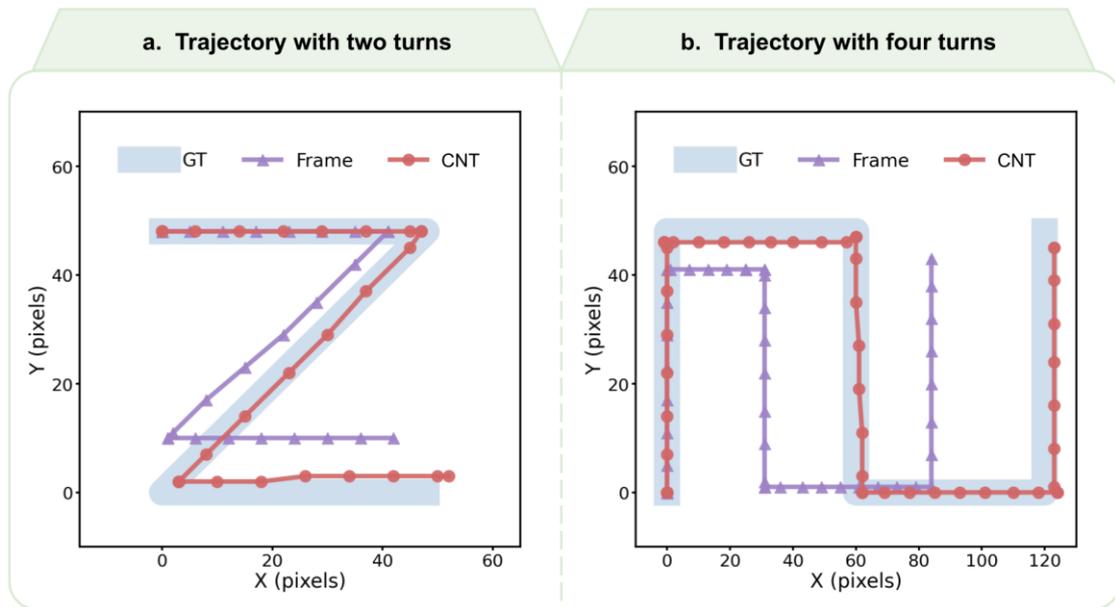

**Fig. 7 Nonlinear trajectory reconstruction.** Two complex nonlinear trajectories at a speed of 1 mm/s and under an illumination of 10.5 lux. (a) A combined lateral and diagonal trajectory. (b) A multi-turn trajectory. Ground truth paths are shown as blue solid lines. Frame-based estimates are shown as purple triangles, while CNT estimates are shown as red circles. CNT accurately reconstructs both trajectories, while the frame-based method shows significant lag and deviations, particularly at turning points.

We compare CNT and frame-based tracking on representative trajectories at 1 mm/s and 10.5 lux in Figure 7. Both methods track initial linear segments accurately. However, during abrupt directional changes, frame-based tracking exhibits increasing lag and deviation from the ground truth, consistent with temporal misalignment and accumulated error. In contrast, CNT maintains stable alignment throughout rapid directional changes. This robustness arises from consistent speckle correlation enabled by adaptive temporal sampling and structured representation, allowing reliable tracking under non-smooth motion. A video demonstration of reconstructing trajectory (a) is shown in supplementary material.

## 3 Discussion

We develop computational neuromorphic tracking (CNT) as a framework that integrates asynchronous sensing with physics-informed speckle analysis. Rather than relying on fixed-exposure acquisition settings or heuristic parameter selection, CNT formulates motion tracking as a joint optimization of spatiotemporal representation. This perspective addresses the challenge of preserving both temporal dynamics and speckle structure under scattering and photon-limited conditions.

The experimental results (Sections 2.2-2.7) demonstrate that CNT extends reliable tracking to 20 mm/s and 0.9 lux, beyond the regime accessible to frame-based methods. The agreement between parameters that maximize $\mathcal{J}(n,\omega)$ and those that minimize tracking error (Section 2.4) suggests that the objective captures key factors governing speckle correlation stability, including contrast, structural integrity, and temporal consistency.

A key observation is that sensing alone is insufficient for robust tracking in scattering environments. Frame-based imaging loses temporal information through integration, while raw event streams remain unstable without structured aggregation (Section 2.2). CNT addresses this by explicitly controlling temporal sampling ($n$) and spatial bandwidth ($\omega$), enabling a representation in which motion-induced speckle evolution remains observable.

The dependence of optimal parameters on motion speed (Section 2.3) highlights the adaptive nature of the framework. Faster motion requires finer temporal sampling, whereas slower motion benefits from increased integration. By enabling post-acquisition adjustment of these parameters, CNT reduces reliance on fixed hardware settings and allows adaptation across varying conditions.

The speed boundary analysis (Section 2.5) shows that CNT maintains low error up to 20 mm/s, while frame-based systems saturate above the single-pixel blur limit. Under low illumination (Section 2.6), CNT preserves temporal information through asynchronous sampling, substantially extending illumination tolerance. For complex trajectories (Section 2.7), CNT maintains stable correlation due to reduced reliance on precise spatial filtering and adaptive temporal integration.

Despite these advantages, several limitations remain. The current study focuses on single-object tracking in static scattering media. Extensions to multi-object scenarios or dynamically evolving media may introduce ambiguity and decorrelation effects [6]. In addition, performance remains constrained by the object size and assumptions such as far-field detection [28].

Future work may explore predictive filtering, adaptive parameter updates, and tighter hardware–algorithm co-design. More broadly, CNT suggests a general approach to constructing task-relevant representations from sparse, asynchronous measurements, which may extend to other computational imaging problems in complex media, including non-line-of-sight imaging [29], flow estimation [30, 31], and deep tissue imaging [32]. The success of recent learning-based approaches in scattering imaging [7, 33, 34] further underscores the potential of integrating physics-informed representations with data-driven strategies.

## 4 Materials and methods

### 4.1 Forward model for neuromorphic speckle formation

Event generation follows a threshold-crossing model [35]. An event $\boldsymbol{e_i} = (\boldsymbol{r}_i, t_i, p_i)$ is triggered at pixel $\boldsymbol{r}_i$ and time stamp $t_i$, when the logarithmic intensity $L(\boldsymbol{r}, t) = \log I(\boldsymbol{r}, t)$ satisfies

$$L(\boldsymbol{r}_i, t_i) - L(\boldsymbol{r}_i, t_i - \Delta t_i) = p_i D, \tag{3}$$

where $p_i \in \{+1, -1\}$ is the polarity and $\Delta t_i$ is the time since the previous event at $\boldsymbol{r}_i$. For sufficiently small $\Delta t_i$,

$$\frac{\partial L}{\partial t}(\boldsymbol{r}_i, t_i) \approx \frac{p_i D}{\Delta t_i}, \tag{4}$$

indicating that events sample the temporal gradients of speckle fluctuations. A detailed derivation is provided in Supplementary Note 4. This mechanism operates without exposure-based integration [35].

To construct displacement-sensitive speckle maps, events are aggregated within a temporal window $\Delta t$ to form a neuromorphic speckle map

$$S(\boldsymbol{r}, t) = \sum_i p_i \, \delta(\boldsymbol{r} - \boldsymbol{r}_i) \, \text{rect}\left(\frac{t_i - t}{\Delta t}\right), \tag{5}$$

where $\delta(\cdot)$ is the Kronecker delta and $\text{rect}(\cdot)$ selects events within $[t - \frac{\Delta t}{2}, t + \frac{\Delta t}{2}]$ (see Supplementary Note 5). The window $\Delta t$ governs a trade-off: shorter windows preserve rapid dynamics, whereas longer windows improve SNR through event accumulation.

In practice, the measured map is corrupted by spurious events unrelated to speckle evolution, such as photon shot noise under low illumination [36]. We model the observation as

$$S_\text{m}(\boldsymbol{r}, t) = S(\boldsymbol{r}, t) + \eta(\boldsymbol{r}, t), \tag{6}$$

where $\eta(\boldsymbol{r}, t)$ is a zero-mean random field with finite variance.

## 4.2 Spatiotemporal representation

Neuromorphic sensing allows the aggregation window to be selected post-acquisition. For consistent comparison with frame-based imaging, we define the dimensionless parameter $n = \tau/\Delta t$, where $\tau$ is the exposure time of a reference camera ($\tau = 40$ ms in our experiments). The aggregated map becomes

$$S_\mathrm{m}(\boldsymbol{r}, t; n) = \sum_i p_i \, \delta(\boldsymbol{r} - \boldsymbol{r}_i) \operatorname{rect}\left(\frac{n(t_i - t)}{\tau}\right) + \eta(\boldsymbol{r}, t; n), \tag{7}$$

with $\eta(\boldsymbol{r}, t; n)$ the temporally accumulated noise. The parameter $n$ controls the balance between temporal decorrelation and event statistics.

To suppress high-frequency noise, we apply a spatial denoising operator $\mathcal{D}_\omega\{\cdot\}$ parameterized by the cutoff frequency $\omega$:

$$S_\mathrm{d}(\boldsymbol{r}, t; n, \omega) = \mathcal{D}_\omega\{S_\mathrm{m}(\boldsymbol{r}, t; n)\}. \tag{8}$$

We adopt a Fourier-transform-based lowpass filter [37, 38] with transfer function $H(\boldsymbol{f}; \omega)$ for its simplicity and explicit frequency control. The denoised map is therefore

$$S_\mathrm{d}(\boldsymbol{r}, t; n, \omega) = \mathcal{F}^{-1}\left\{H(\boldsymbol{f}; \omega) \mathcal{F}\left\{\sum_i p_i \, \delta(\boldsymbol{r} - \boldsymbol{r}_i) \operatorname{rect}\left(\frac{n(t_i - t)}{\tau}\right) + \eta(\boldsymbol{r}, t; n)\right\}\right\}. \tag{9}$$

This formulation separates temporal aggregation and spatial filtering, enabling joint control of spatiotemporal representation via the parameter pair $(n, \omega)$.

## 4.3 Physics-informed speckle analysis

Conventional metrics such as global speckle contrast do not align well with tracking performance, as they may favor noisy, high-contrast patterns (see Supplementary Note 6). We instead define a physics-informed objective function tailored to correlation-based motion estimation.

The average local speckle contrast over sliding neighborhoods is

$$K = \mathbb{E}_{\boldsymbol{r}}\left[\frac{\sigma_{S_\mathrm{d}(\boldsymbol{r})}}{|\mu_{S_\mathrm{d}(\boldsymbol{r})}| + \varepsilon}\right], \tag{10}$$

which emphasizes local fluctuations that contribute to correlation peak sharpness. To prevent over-smoothing and preserve the structural fingerprint of speckle patterns, we include the spatial gradient energy

$$G = \langle |\nabla S_\text{d}(\boldsymbol{r}, t)|^2 \rangle. \tag{11}$$

Under the small-displacement regime ensured by operation within the optical memory effect, temporal correlation serves as a reliable proxy for inter-frame similarity:

$$M = \frac{\sum_r [(S_\text{d}(\boldsymbol{r}, t) - \mu_t)(S_\text{d}(\boldsymbol{r}, t + \Delta t) - \mu_{t+\Delta t})]}{\sqrt{\sum_r (S_\text{d}(\boldsymbol{r}, t) - \mu_t)^2} \sqrt{\sum_r (S_\text{d}(\boldsymbol{r}, t + \Delta t) - \mu_{t+\Delta t})^2}}, \tag{12}$$

where $\mu_t = \langle S_\text{d}(\boldsymbol{r}, t) \rangle_r$.

These terms are combined into a physics-driven objective

$$\mathcal{J}(n, \omega) = \alpha K + \beta G + \gamma M, \tag{13}$$

with empirically determined weights $\alpha = 1$, $\beta = 0.2$, and $\gamma = 5$, held fixed across all experiments. The optimal spatiotemporal parameter pair is

$$(n^*, \omega^*) = \underset{n \in N,\ \omega \in \Omega}{\text{argmax}}\ \mathcal{J}(n, \omega), \tag{14}$$

which is optimized for motion observability and stable displacement inference, rather than visual contrast. We solve the optimization problem over integer sets $N$ and $\Omega$, where $N = \{1, \ldots, 10\}$ and $\Omega = \{20, \ldots, 60\}$, constrained by the physical limits of the sensing system. For each $(n, \omega)$ pair, we compute the denoised map and the objective $\mathcal{J}$. A coarse grid search over $N \times \Omega$ identifies the region of maximal $\mathcal{J}$, followed by a coordinate-wise refinement to converge to $(n^*, \omega^*)$. The optimal parameters are then applied to form the final speckle maps for motion estimation.

### 4.4 Calibration-free motion estimator

With the optimized parameter pair $(n^*, \omega^*)$, the maps $S_\text{d}(\boldsymbol{r}, t)$ encode displacement-sensitive features. Motion is estimated using speckle correlography, which relies on the intrinsic spatial correlation of speckle patterns and requires no prior knowledge of the scattering medium or motion model [2].

To mitigate spatial biases arising from non-ideal illumination and collection geometry [39], we first subtract the spatial mean:

$$\tilde{S}(\boldsymbol{r},t) = S_d(\boldsymbol{r},t) - \langle S_d(\boldsymbol{r},t)\rangle. \tag{15}$$

The displacement between two frames $t_i$ and $t_j$ is then inferred from the peak of the mean-subtracted cross-correlation

$$\tilde{C}_{i,j}(\boldsymbol{r}) = \tilde{S}(\boldsymbol{r},t_i) \star \tilde{S}(\boldsymbol{r},t_j), \tag{16}$$

relative to $\tilde{C}_{i,i}(\boldsymbol{r})$. This mean-subtracted formulation improves robustness without calibration or prior assumptions (see Supplementary Note 7).

## 4.5 Recursive tracking beyond OME

The tracking range of speckle correlography is fundamentally constrained by OME, which limits the maximum displacement $d_{\text{OME}}$ over which speckle patterns remain correlated [2]. Beyond this limit, direct correlation between distant frames fails.

CNT enables reliable estimation of small successive displacements, which allows the reference to be updated sequentially. Consecutive maps $\tilde{S}_i$ and $\tilde{S}_{i+1}$ are constructed such that $\Delta \boldsymbol{d}_{i\to i+1}$ remains within the OME regime. After each step, the reference is updated ($\tilde{S}_i \leftarrow \tilde{S}_{i+1}$), and displacements are accumulated:

$$\Delta \boldsymbol{d}_{1\to n} = \sum_{k=0}^{n-1} \Delta \boldsymbol{d}_{k\to k+1}. \tag{17}$$

The cumulative displacement can thus exceed the intrinsic memory effect limit:

$$\| \Delta \boldsymbol{d}_{1\to n} \| \gg d_{\text{OME}}. \tag{18}$$

This strategy preserves the validity of each local correlation while extending the effective tracking range through sequential accumulation. Combined with high-temporal-resolution neuromorphic sensing, it enables large-scale motion tracking in scattering media without prior knowledge of system parameters.

## 4.6 Data acquisition

The experimental setup is depicted in Figure 1b. A continuous-wave 532 nm laser (200 mW DPSS) is expanded by a $2 \times$ beam expander (BE) and passed through a rotating diffuser (RD) to reduce spatial coherence, suppressing intense speckle hotspots and improving measurement stability [40]. The beam is then collimated and further expanded by a $10 \times$ BE before illuminating a negative USAF resolution target (Thorlabs R3L3S1N) mounted on a motorized translation stage, allowing controlled two-dimensional displacement. The element '5' of group 0 is filtered to transmit light. Light passing through the moving target then traverses a static 220-grit ground-glass diffuser (Edmund Optics), generating dynamic speckle patterns. An iris between the diffuser and camera reduces stray light to improve contrast. See Supplementary Note 8 for the real optical experimental setup.

Detection is performed by an event camera (Inivation DAVIS346, $346 \times 260$ pixels) equipped with a $20 \times$ objective lens to magnify the speckle pattern rather than form a focused image. The camera outputs 25 fps intensity frames and asynchronous event streams. Distances from the target to the diffuser and from the diffuser to the sensor image plane are $u$ and $v$, respectively. We calibrate the relationship between physical velocity and pixel displacement with $u = 125$ mm and $v = 125$ mm. Controlled experiments show that an object speed of 1 mm/s corresponds to approximately 7.5 pixels per second on the sensor, enabling direct conversion between pixel shifts and physical speeds.

## Supplementary information

See Supplementary Information and Supplementary Video for supporting content.

## Declarations

- Funding: This work is supported by Theme-based Research Scheme (T45-701/22-R) and Research Grants Council of Hong Kong (GRF17201822).

- Competing interests: The authors declare no conflicts of interest.

- Availability of data and materials: Data underlying the results presented in this paper are not publicly available at this time but may be obtained from the authors upon reasonable request.

- Authors' contributions: Yuqing Cao, Shuo Zhu, and Edmund Y. Lam conceived and designed the study. Yuqing Cao, Shuo Zhu, and Rongzhou Chen developed the experimental setup and performed the data acquisition. Yuqing Cao implemented the computational algorithms and carried out data analysis. Yuqing Cao, Shuo Zhu, Jingyan Chen, and Ni Chen interpreted the results. Edmund Y. Lam provided supervision throughout the project. Yuqing Cao wrote the initial draft of the manuscript. All authors participated in discussions and commented on the manuscript.

- Acknowledgements: We are grateful to all members of the Imaging Systems Laboratory in HKU for their support and valuable suggestions.